\documentclass[a4paper, 10pt, conference]{ieeeconf}      
\usepackage{FG2019}

\FGfinalcopy 

\IEEEoverridecommandlockouts                              
 \usepackage{amsmath} 
\usepackage{graphicx}

\usepackage{float}

\def\FGPaperID{215} 

\title{\LARGE \bf
3D Feature Pyramid Attention Module for Robust Visual Speech Recognition
}

\begin{document}

\ifFGfinal
\thispagestyle{empty}
\pagestyle{empty}
\author{\parbox{16cm}{\centering
    {\large Jingyun Xiao$^{1,2}$}\\
    {\normalsize
    $^1$ Department of Computer Science, University of Chinese Academy of Sciences, Beijing, China\\
    $^2$ Key Laboratory of Intelligent Information Processing of Chinese Academy of Sciences (CAS), Institute of Computing Technology, CAS, Beijing, China
    }}
}
\else
\author{Anonymous FG 2019 submission\\ Paper ID \FGPaperID \\}
\pagestyle{plain}
\fi
\maketitle

\begin{abstract}

Visual speech recognition is the task to decode the speech content from a video based on visual information, especially the movements of lips. It is also referenced as lipreading. Motivated by two problems existing in lipreading, words with similar pronunciation and the variation of word duration, we propose a novel 3D Feature Pyramid Attention (3D-FPA) module to jointly improve the representation power of features in both the spatial and temporal domains. 
Specifically, the input features are downsampled for $3$ times in both the spatial and temporal dimensions to construct spatiotemporal feature pyramids. Then high-level features are upsampled and combined with low-level features, finally generating a pixel-level soft attention mask to be multiplied with the input features.
It enhances the discriminative power of features and exploits the temporal multi-scale information while decoding the visual speeches. Also, this module provides a new method to construct and utilize temporal pyramid structures in video analysis tasks. The field of temporal featrue pyramids are still under exploring compared to the plentiful works on spatial feature pyramids for image analysis tasks. 
To validate the effectiveness and adaptability of our proposed module, we embed the module in a sentence-level lipreading model, LipNet \cite{Assael2016}, with the result of $3.6\%$ absolute decrease in word error rate, and a word-level model proposed in \cite{Stafylakis2017}, with the result of $1.4\%$ absolute improvement in accuracy.

\end{abstract}

\section{Introduction}

Visual speech recognition, also known as lipreading, is a developing topic in the field of video understanding which receives more and more attention in recent years. 
It has broad application prospects in hearing aids and special education for hearing impaired people, complementing speech recognition in noisy environments, new man-machine interaction methods and other potential application scenarios.

The target of the lipreading task is to decode the speech content from videos based on the visual information, including the how the lips, tongue, teeth move and interact during the speaking process. Different words have different pronunciation corresponding to different lip-region movements, therefore it is possible to learn the consistent latent patterns of the same words spoken by different speakers and to discriminate different words based on these patterns.

\begin{figure}[H]
  \centering
  \includegraphics[width=1.0\linewidth]{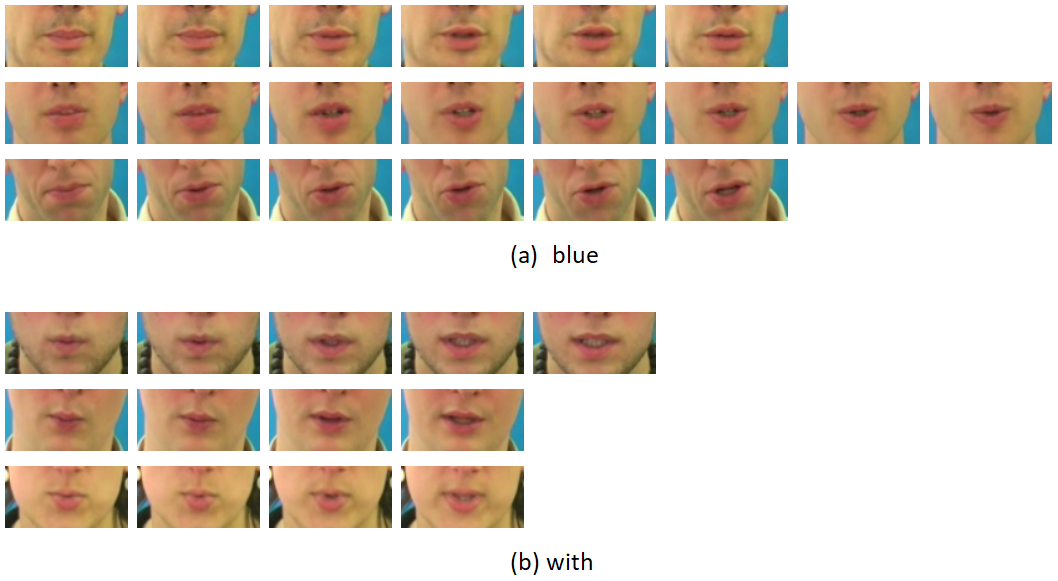}
  \caption{The same words spoken by different people. The target of lipreading is  to exploit the latent patterns of the image sequences of the same words. }
  \label{fig:blue_with}
\end{figure}

\begin{figure*}[htbp]
  \centering
  \includegraphics[width=1.0\linewidth]{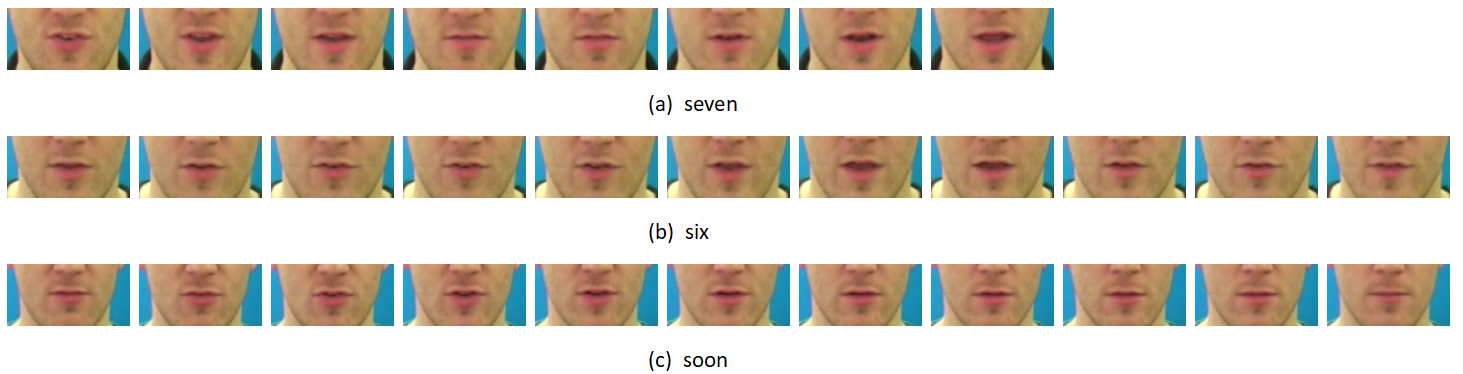}
  \caption{Different words spoken by the same person. There are tiny differences between frames of different words, which is a common case in the lipreading task.  }
  \label{fig:seven_six_soon}
\end{figure*}

Figure \ref{fig:blue_with} shows the lip crop sequences of the same word spoken by different speakers, indicating a consistent visual pattern that makes the sequence to express the certain word ``blue'' or ``with'' instead of ``red'' or ``at''. Figure \ref{fig:seven_six_soon} shows different words spoken by the same person. As shown, only tiny differences exist in the initial or terminal frames of different words, which is a common case in the lipreading task.

\begin{figure}[H]
  \centering
  \includegraphics[width=1.0\linewidth]{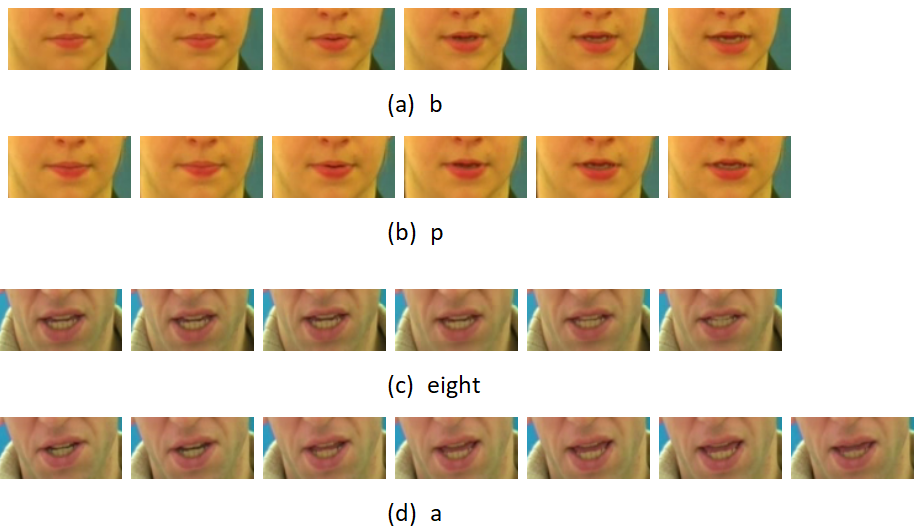}
  \caption{Words with similar pronunciations are difficult to distinguish.}
  \label{fig:b_p_eight_a}
\end{figure}

Lipreading has common features and processing steps with other video tasks such as activity detection. Moreover, it has its own characteristics and faces specific difficulties. One major difficulty is how to distinguish the words with similar pronunciation.  As shown in Figure \ref{fig:b_p_eight_a}, it is difficult to distinguish words with similar pronunciation, especially for short words. To address this problem, we expect to enhance the discriminative power of spatial features on each frame. Another challenge is that many words have a short duration, usually no more than 0.02 seconds, therefore unable to provide enough information to learn the latent patterns. In our experiments, short words such as ``a'', ``an'', ``eight'' and ``bin'' suffer higher error rates. To addres this problem, the temporal context information should be referred while decoding the words. Moreover, a good model is expected to recognize words at different temporal scales, which requires temporal multi-scale information.

Motivated by the two problems above, we expect a method to enhance the spatial features in each frame as well as explore the multi-scale information in the temporal dimension of the whole video. Multi-scale problem has always been a concern in computer vision researches. In image processing tasks such as object detection and semantic segmentation, plentiful methods are proposed to detect the target objects at different scales. Feature Pyramid Network \cite{Lin2017} and related methods have shown great effect on this problem. Recently \cite{Li2018} proposes a Feature Pyramid Attention module which combines attention mechanism with spatial pyramid in the semantic segmentation task. It refers local and global information at different level to enhance the localization information of features, improving the prediction accuracy especially on small objects, which exactly coincides with our requirements. We adapt the FPA module in \cite{Li2018} to the lipreading task, generalize it to the temporal dimension and propose a novel 3-Dimension Feature Pyramid Attention (3D-FPA) module to enhance the discriminative power on each frame and exploits the temporal multi-scale information of the whole video. The core point is to utilize high-level features as guidance to the attention of low-level features and generate a pixel-level attention mask on the input features. This method is inspired by FPN \cite{Lin2017} and SENet \cite{Hu2017}. 

The main contributions of this paper are summarized as follows. (1) We propose a 3D Feature Pyramid Attention (3D-FPA) module to improve the representation power of spatiotemporal features for the lipreading task. (2) Our proposed model provides a new method to construct and utilize temporal pyramid structure, which has referential value for other video analysis tasks. (3) We demonstrate the effectiveness and adaptability of our module by embedding it into a sentence-level prediction model, LipNet \cite{Assael2016} and a word-level classification model, CRNL \cite{Stafylakis2017} and performing experiments on the GRID corpus \cite{Cooke2006} and the LRW corpus \cite{Chung2016} respectively.

The rest of this paper is organized as follows. In Section 2, we review recent works on feature pyramids and lipreading. In Section 3, we present the details the proposed module and the modified architectures of two lipreading models. In Section 4, we evaluate the models on two large-scale database and provide a detailed analysis of the experiment results. Finally, we make conclusion of our work in section 5. 

\section{Related Work}

In this section, we summarize our literature review on works of feature pyramids and approaches over lipreading. In the first part, we review the works of spatial pyramid which is widely used in image processing tasks. Then we outline recent works on temporal pyramid used in video analysis tasks. In the last part, we give a basic review over the existing methods on lip reading.

\subsection{Spatial Pyramid}

In the object detection task, detecting and recognizing objects at various scales has always been a concern. In the era of hand-engineered features, the original image is resized to a range of scales and constructs a pyramidal structure that is called image pyramids. Models detect objects on each level of the image pyramids.

With the advance of deep learning methods, different approaches have been proposed to solve the multi-scale problem, among which two categories of approaches are popular. The first category performs detectors of different scales on the same feature map extracted by neural networks.  Fast RCNN \cite{Girshick2015} and Faster RCNN \cite{Ren2015} use anchors of 3 different sizes and 3 length-width ratios, 9 anchors in all, to detect objects at different scales. The second category utilizes the inherent pyramidal structure of featrues maps produced by different layers of deep neural networks in the forward  propagation, which is called feature pyramids, and perform detection on different levels of the pyramids with a shared detector. 

One representative work is Feature Pyramid Networks (FPN) \cite{Lin2017}, which provides a novel and efficient way to build feature pyramids with high-level semantic information at all levels by combining low-level, semantically weak features with high-level, semantically strong features via a top-down pathway and lateral connections. Compared to the methods of the first category, methods of feature pyramids provide the multi-scale information with marginal time and memory extra cost because the models reuse the feature maps which are computed in the forward propagation. Moreover, feature pyramids can be utilized in other image analysis tasks. 

Since feature pyramids have achieved appealing results in the task of object detection, pyramidal structure is introduced to scene parsing and semantic segmentation to explore the multi-scale information of the image. PSPNet \cite{Zhao2012} upsamples feature pyramids of different levels to a certain size and then concatenate all the upsampled features together to get a finest feature map rich in both local and global context information. 

Inspired by FPN \cite{Lin2017} and SENet \cite{Hu2017}, Hanchao Li et. al \cite{Li2018} propose a Feature Pyramid Attention (FPA) module to combine feature maps at different levels by upsampling and lateral connections, finally generating a pixel-level soft attention mask to be multiplied with the input feature maps. It combines attention mechanism and spatial pyramids to enhance important features based on the reference of local and global context information. It shows great performance in the semantic segmentation task especially for small objects. We modify the FPA module and generalize it from 2D to 3D to get better representation in both spatial and temporal domains.

\subsection{Temporal Pyramid}

Inspired by the success of spatial pyramid structures in object detection, a few recent works attempt to apply pyramidal structures in the temporal dimension to activity detection, expecting to perform better detection on activity instances over different temporal scales. Temporal pyramid structures have shown its potential power in exploiting the multi-scale information of videos. However, since video analysis is much more complicated than image analysis, how to design effective temporal pyramid architectures and combine them with the backbone networks properly is still under exploring, which is a main concern of our work in this paper. In this part, we would give a general review of three recent works that employ temporal pyramid structures in different manners for the task of  activity detection .

Zhang et al. \cite{Zhang2018} propose Dynamic Temporal Pyramid Network (DTPN) which samples an input image sequence at different frequency to generate multiple image sequences with different temporal sizes, thus constructing pyramidal input data. The pyramidal input is passed forward to the following modules to extract features independently to build feature pyramids. The feature maps at different levels are upsampled to the same scale and concatenated together to combine the local and global information. DTPN achieves state-of-the-art results on the ActivityNet dataset. This practice resembles the method of image pyramids in early times for object detection, which resizes the input image to different scales to build pyramidal structures. However, building feature pyramids from pyramidal input causes high computation cost. 

Contextual Multi-Scale Region Convolutional 3D Network (CMS-RC3D) \cite{Bai2018} uses the C3D ConvNet as the backbone architecture to extract features and add two additional downsampling layers to construct temporal feature pyramids. On each level of the temporal feature pyramids, an activity proposal detector and an activity classifier are learned to detect activities of specific temporal scales independently. Compared to \cite{Zhang2018}, this method builds feature pyramids by reusing feature maps computed by previous layers, therefore reduces computation cost. It does not fuse the features at different levels of the pyramids. It performs detection and classification on each level of pyramids independently without fusing the multi-scale information. 


Structured Segment Networks (SSN) \cite{Zhao2017} proposes structured temporal pyramid pooling (STPP) to produce a global representation of the each generated proposal. For each proposal, it extracts spatiotemporal features, builds 2-level feature pyramids and concatenates them to get global representation for the proposal, which is used for the following activity and completeness classification. Combining multi-scale information of a proposal and its surrounding snippets brings a good balance between expressive power and complexity, since it does not build feature pyramids on the whole video. 

The methods above work well for the activity detection task but has limited instructive value for other video analysis tasks such as lipreading and video description, which require refined and consecutive spatiotemporal information.

In our proposed method, we build feature pyramids by reusing feature maps computed in previous layers, therefore it requires marginal extra computation cost. Since the lipreading task is a decoding task but not a detection task, we build pyramids on the whole video rather than focus on several snippets of  the video to provdie intact information of the whole sequence. Also, in order not to lose temporal distribution information, we do not concatenate features together but generate a pixel-wise attention mask on the original feature maps to select import featrues, inspired by the chanel-wise mask of SENet \cite{Hu2017}.

\subsection{Lipreading}

\begin{figure}
  \centering
  \includegraphics[width=1.0\linewidth]{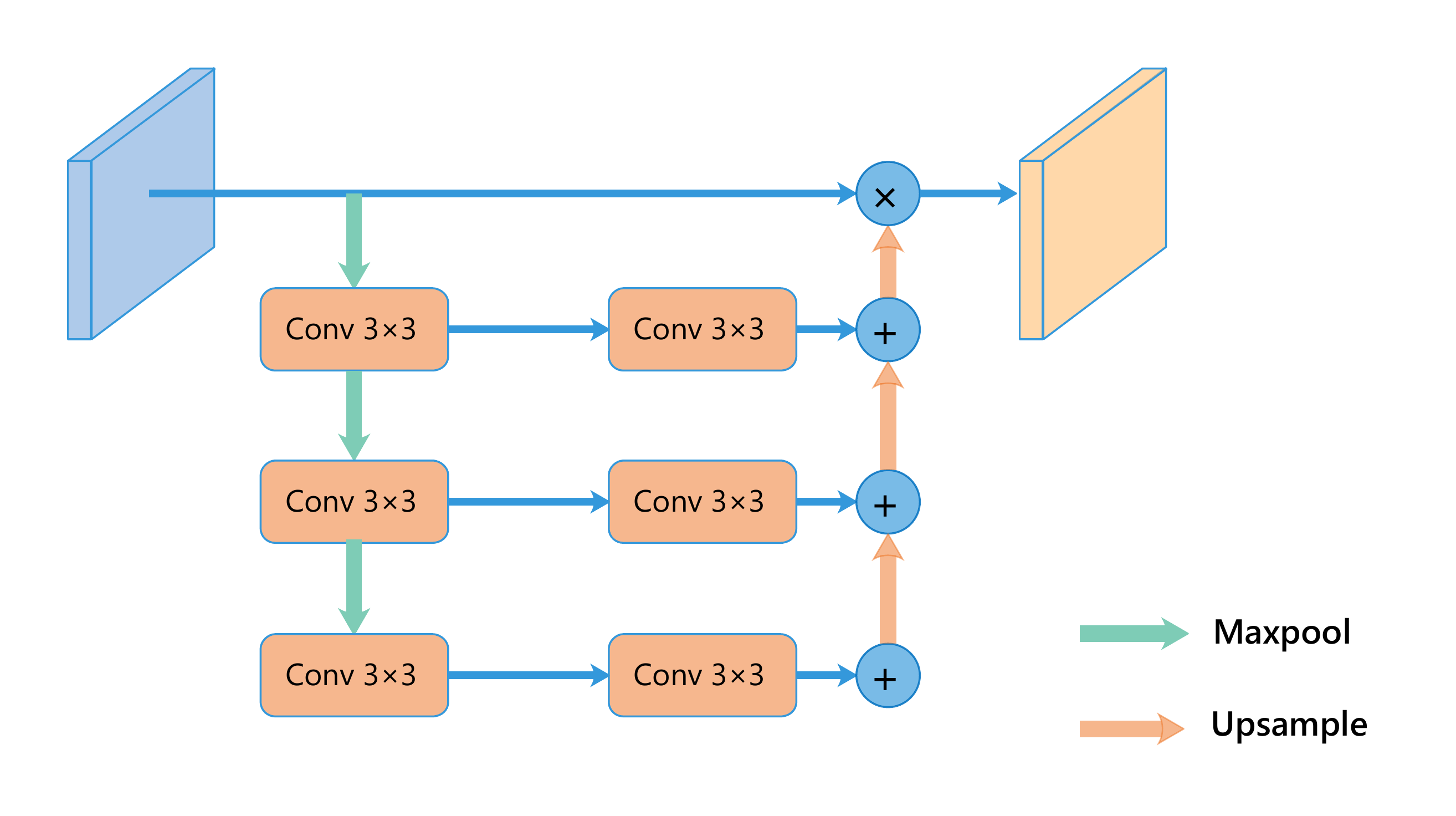}
  \caption{The structure of 2D-FPA module we adopt for lipreading. The green and orange arrows denote the downsampling and upsampling operations respectively. This module utilizes high-level features as guidance to low-level features, therefore combines information at different levels step by step. Finally it generates an attention mask for the input feature maps.}
  \label{fig:2d_fpa}
\end{figure}

In the early stage, most of the works on visual speech recognition focused on how to design proper feature extractors to represent the complicated lip movement sequences. A classical process is to employ Hidden Markov Models to exploit the temporal relationships among the extracted features \cite{Goldschen1997, Chiou1997, Potamianos2003,Chandrasekaran2009}. With  the development of deep learning technologies and the appearance of large-scale lipreading databases, a few works started to introduce the convolutional neural network to extract the features of each frame and employ recurrent units to model the temporal relationships of the frame sequence in the speaking process \cite{Noda2015, Thangthai2015, Almajai2016}. In 2016, \cite{Chung2016} proposed the first end-to-end lipreading model which performs word-level classification as well as a large-scale word-level database LRW. After that, more recent lipreading approaches follow an end-to-end fashion. 

Based on the observation that human lipreaders perform better on long words an sentences, \cite{Assael2016} proposed LipNet, the first end-to-end lipreading model that performs sentence-level prediction. It takes frame sequences of  variable length as input and outputs character sequences. It uses three cascaded spatiotemporal convolutional neural networks to extract spatiotemporal features and recurrent units to perform character prediction at each time-step. Notably, it employs the CTC loss \cite{Graves2012}, which is widely used in the speech recognition task to train unaligned datas. It attains the word error rate equal to $11.4\%$ on the sentence-level database GRID \cite{Cooke2006}. Compared to word-level models, LipNet can exploit temporal context when predicting sentences, therefore attains much  higher accuracy. The result accords with the fact that human lipreaders perform better on longer words or sentences. The detailed architecture of LipNet is presented in section 3. 

Then In 2017, \cite{Stafylakis2017} proposed a complicated word-level model and attained word accuracy equal to $83.0\%$ on the LRW corpus, making the new state-of-the-art result. It is a combination of spatiotemporal convolutional units, ResNet \cite{He2016} and bidirectional LSTM networks \cite{Graves2005}. It introduces ResNet to cope with the massive amount of datas with extraordinarily high variability in the LRW corpus. The detailed architecture is presented in section 3.

As stated in Section 1, lipreading has two difficulties to deal with. The words with similar pronunciations are hard to distinguish and short words are unable to provide enough information for precise recognition. Inspired by the works on spatial pyramid and temporal pyramid, we propose a 3D-FPA module to enhance the discriminative power on each single frame as well as exploit the temporal multi-scale information of the whole frame sequence.

\begin{figure*}
  \centering
  \includegraphics[width=0.9\linewidth]{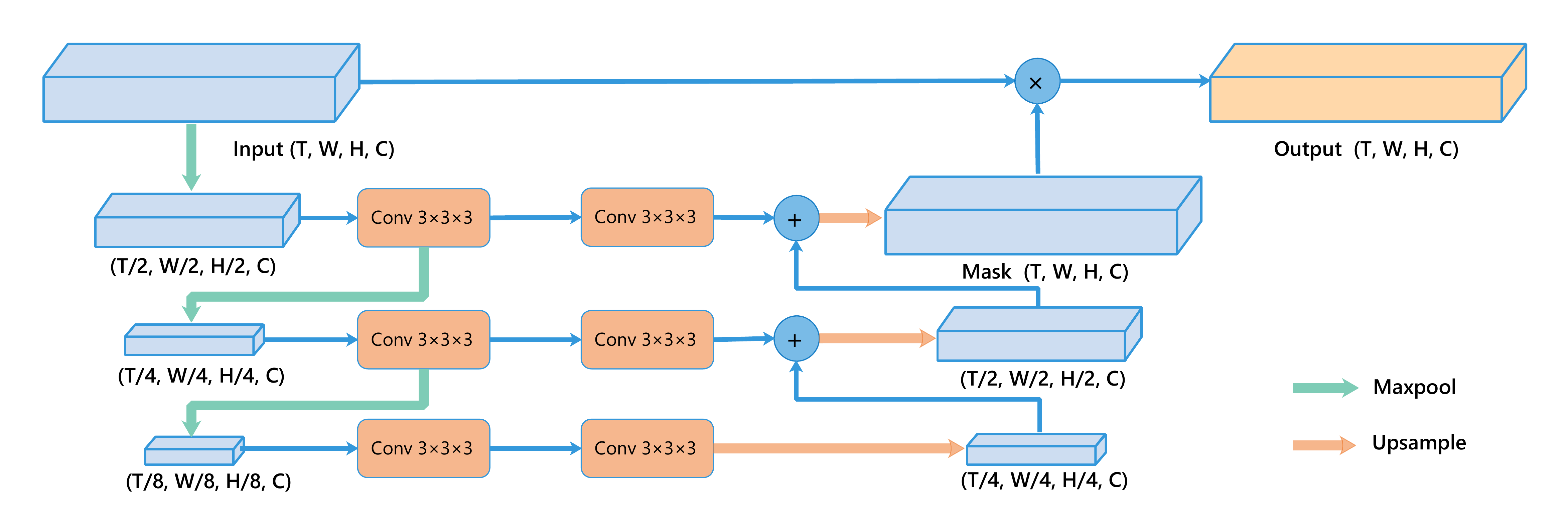}
  \caption{The structure of 3D-FPA module. T, W, H, C denote the size of  the time-step, width, height and channel respectively. It constructs feature pyramids in both spatial and temporal domains simultaneously. High-level features are utilized to guide the attention of low-level features. It exploits the multi-scale information in spatial and temporal dimensions jointly.}
  \label{fig:3d_fpa}
\end{figure*}

\section{Method}

In this section, we first present the 2D-FPA module proposed in \cite{Li2018}. Then we describe the architecture of our 3D-FPA module. After that we introduce two lipreading architectures respectively performing sentence-level prediction and word-level classification and how we embed the FPA module in the architectures to verify the effectiveness and adaptability our proposed module.

\subsection{2D-FPA}

In semantic segmentation, models employing spatial pyramid pooling such as PSPNet \cite{Zhao2012} may lose local information. Inspired by Attention Mechanism and SENet \cite{Hu2017} using channel-wise attention to weight the feature maps, \cite{Li2018} proposes Feature Pyramid Attention (FPA) module that produces pixel-level attention with context  prior information of each pixel to select the features pixel-wisely.

FPA deploys bottom-up and top-down branches similar to Feature Pyramid Network \cite{Lin2017}. In the bottom-up branch, feature maps are extracted at 3 different scales. It use $3\times 3$, $5\times 5$, and $7\times 7$ convolution kernels at different pyramid levels. In the top-down branch, the context information of different scales is integrated step-by-step, from global to local, finally generates an attention mask with the same shape of the original feature map. The original feature map is passed through $1\times 1$ convolution and multiplied pixel-wisely with the attention mask. The author also introduces a global pooling branch which is concatenated with the output features to further improve the performance.

For our lipreading task, we expect to improve the discriminative ability on each single frame to address the problem of different words with similar lip shapes. We employ the FPA module to exploit the nuanced information and focus on important features with the pixel-level attention mask. 

To adapt the FPA module for our task, we do several experiments with different settings and find the best structure. We remove the $1\times 1$ convolution before the multiplication between the mask and the input feature map to protect the input feature map from losing information. We also remove the global pooling branch for lower computation cost. We use $3\times 3$ convolution kernels at all pyramid levels. The final structure we adopt is presented in Figure \ref{fig:2d_fpa}.

\subsection{3D-FPA}

To address the temporal multi-scale problem, we naturally think about temporal pyramid structure. However, improper structures may cause information loss in temporal distribution, which is especially harmful to the performance in lipreading. Inspired by the mechanism of  the FPA module in \cite{Li2018}, generating a soft attention mask with prior information of different temporal scales is a practical solution. We extend the 2D-FPA to the temporal dimension and design a 3D-FPA module for the lipreading task.  

It takes a sequence of feature maps as input and outputs pixel-wisely weighted feature maps with the same shape of the input. The structure is similar to 2D-FPA. It is noteworthy that the temporal dimension is also downsampled and upsampled in the pathway. Figure \ref{fig:3d_fpa} shows the detailed structure of 3D-FPA. Since high-level features have stronger semantic information, it is expected to guide the attention of low-level features, in both temporal and spatial domains simultaneously.

The proposed 3D-FPA module is simple, lightweight and effective. Since its input and output are of the same shape, we can embed it into different backbone architectures. We do most experiments with the LipNet \cite{Assael2016} backbone on the GRID corpus \cite{Cooke2006}, performing sentence-level prediction. And in order to validate the adaptability of 3D-FPA, we also apply it to Combining Residual Networks with LSTMs \cite{Stafylakis2017}, a much more complicated architecture to do word-level classification on the more challenging corpus, LRW \cite{Chung2016}.

\subsection{LipNet}

\begin{figure}[htbp]
  \centering
  \includegraphics[width=0.9\linewidth]{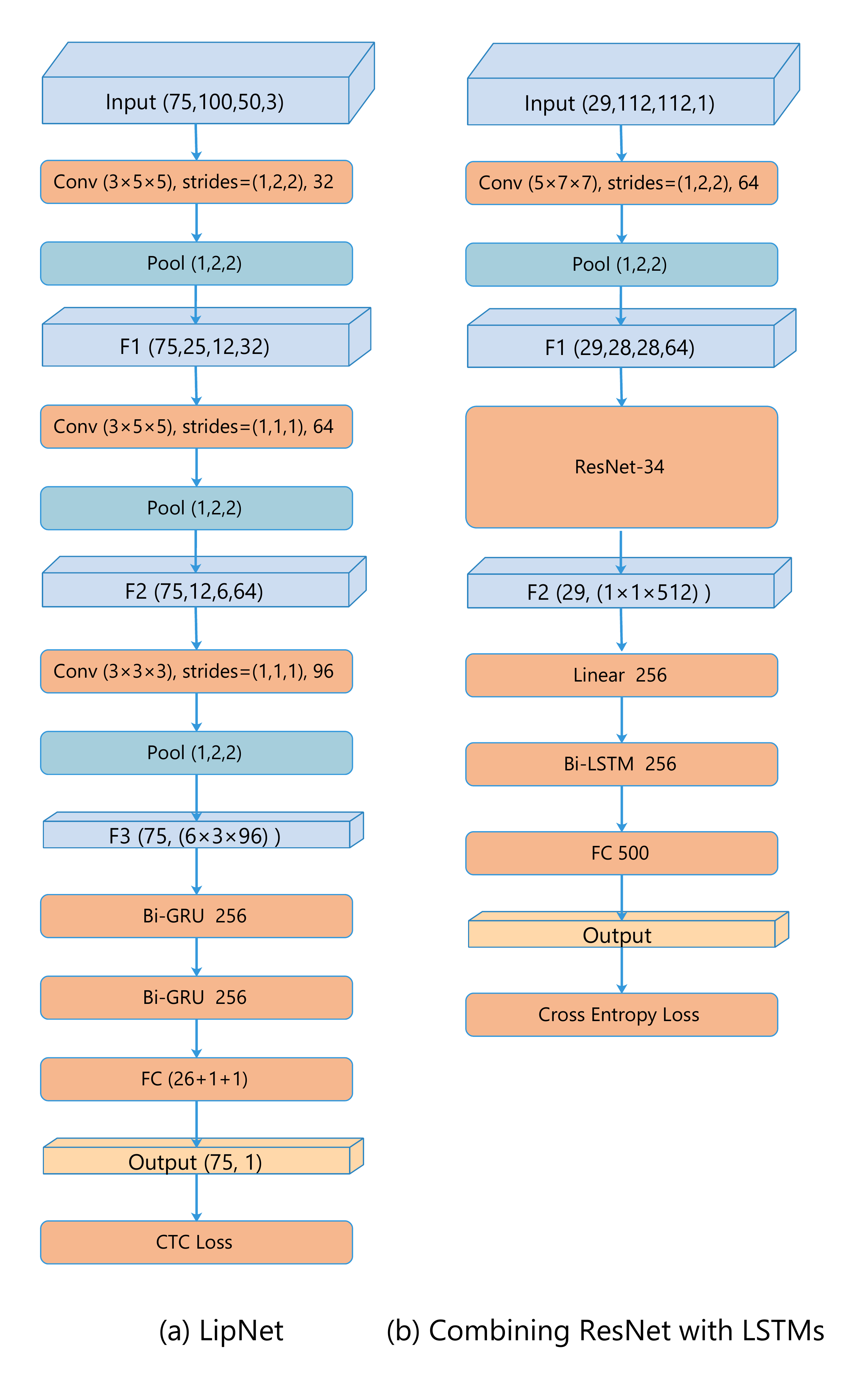}
  \caption{(a) is the structure of LipNet \cite{Assael2016}, which performs sentence-level prediction on the GRID corpus \cite{Cooke2006}. (b) is the structure of CRNL \cite{Stafylakis2017}, which performs word-level classification on the LRW corpus \cite{Chung2016}.}
  \label{fig:lipnet}
\end{figure}

LipNet is the first end-to-end sentence-level lipreading model. It starts with 3 sets of spatiotemporal convolution layers, dropout layers and spatial max-pooling layers. The extracted features are passed forward to two Bi-GRUs. Finally, a linear transformation and a SoftMax are applied at each time-step, followed by the CTC loss. The number of SoftMax output classes of each time-step is 28, 26 letters, a blank symbol and the CTC blank token. The structure of LipNet is illustrated in Figure \ref{fig:lipnet} (a). 
	
\begin{figure}[htbp]
	\centering
	\includegraphics[width=1.0\linewidth]{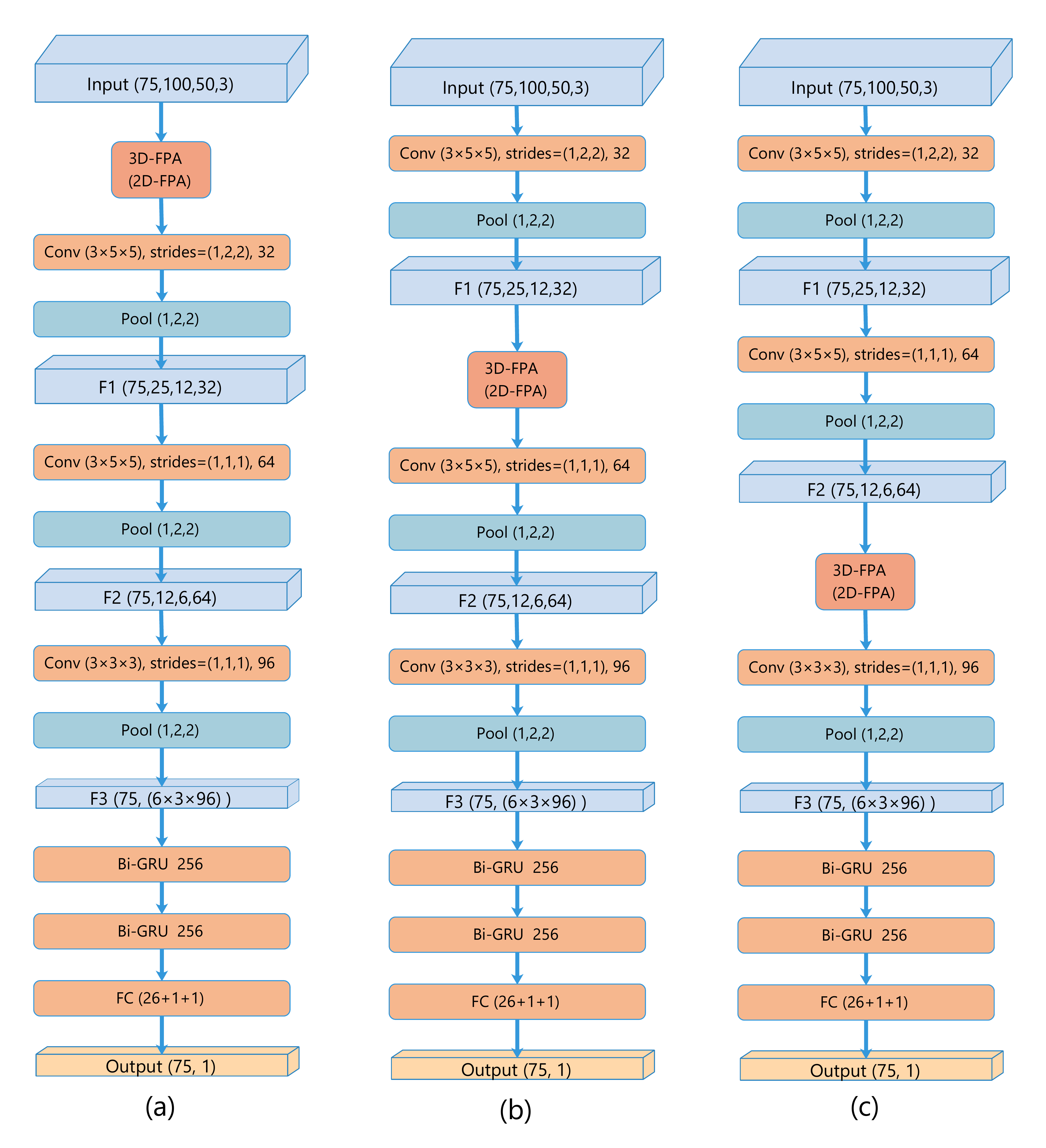}
	\caption{The modified LipNet with FPA modules embedded after the input, F1 and F2 respectively.}
	\label{fig:exp}
\end{figure}

We embed 2D-FPA and 3D-FPA after the input images, F1 and F2 respectively, as shown in Figure \ref{fig:exp} (a) (b) (c). The backbone remains the same with the original architecture of LipNet. 

\subsection{Combining ResNet with LSTMs}

\begin{figure}[htbp]
	\centering
	\includegraphics[width=0.8\linewidth]{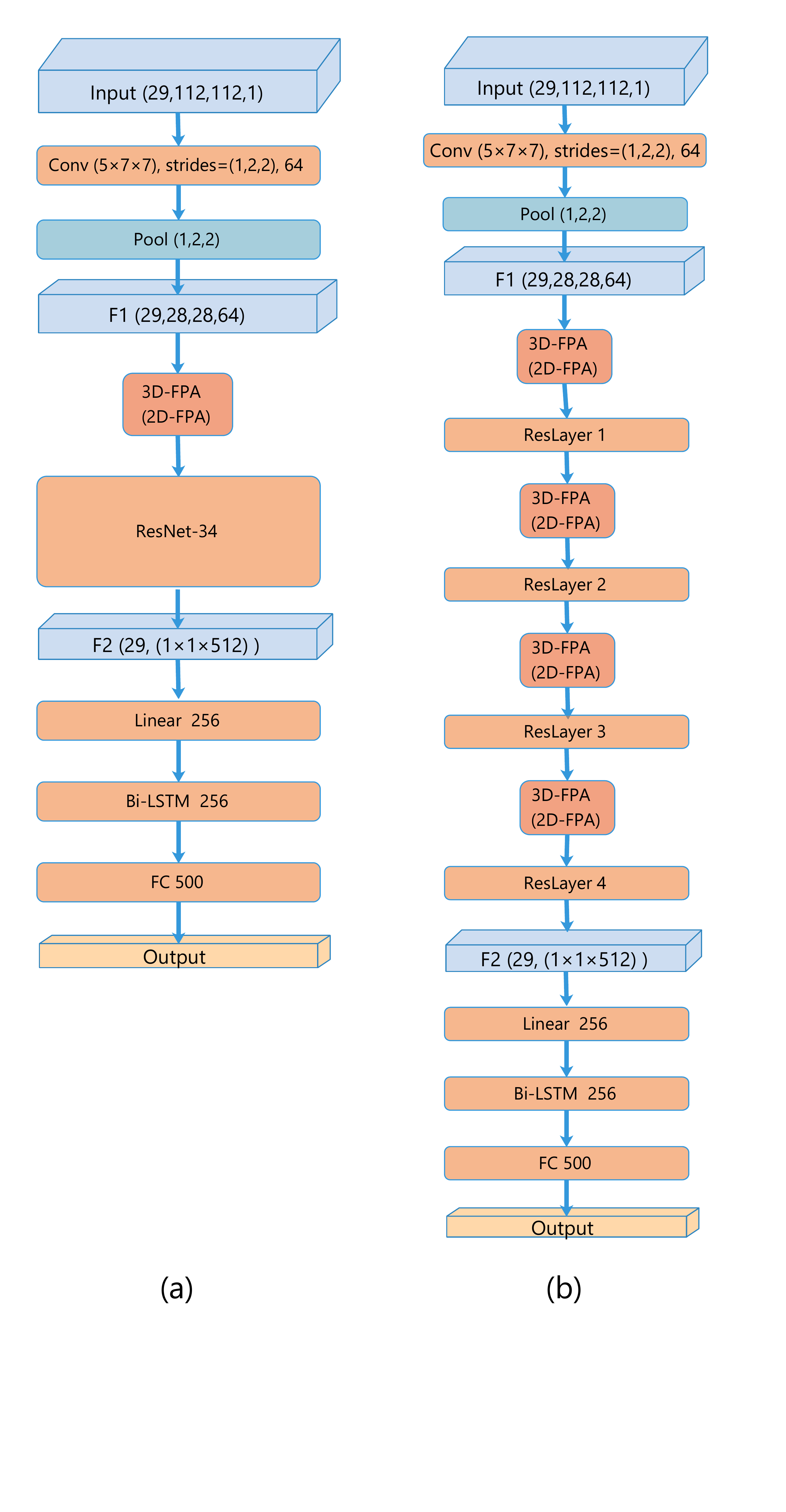}
	\caption{The modified CRNL with FPA modules embedded prior to and inside the ResNet respectively.}
	\label{fig:exp2}
\end{figure}

\cite{Stafylakis2017} proposes a word-level classification model. It starts with a spatiotemporal convolutional front-end, and then a Residual Network is applied to each time-step. Since it is trained and evaluated on LRW, a much more complex database, the ResNet can help to cope with the complicated image features. Then it is followed by a Bidirectional Long Short-Term Mem-ory (Bi-LSTM) network. Finally a SoftMax layer with 500 output classes is followed. The detailed structure is illustrated in Figure \ref{fig:lipnet} (b). 
We apply 2D-FPA and 3D-FPA modules prior to and inside ResNet, as shown in Figure \ref{fig:exp2}. 

\section{Experiments}

We verify the effectiveness and adaptability of the proposed module by embedding  it into the backbone lipreading models and evaluate the modified model on the corresponding datasets. In this section, we present our sentence-level and word-level experiments respectively, including the datasets, implementation details, results and analysis.

\subsection{Sentence-level}

\subsubsection{Dataset}

The GRID corpus is a sentence-level dataset, which contains audio and video recordings  of 34 speakers and each of them produces 1000 sentences. The  total duration of GRID is about 28 hours. The sentences produced with a fixed grammar: command + color + preposition + letter + digit + adverb. The detailed components of the 6 categories are shown in Table \ref{tab:grid}. For example, two sentences in the dataset are ``Bin blue at A zero again'' and ``Place green with B eight soon''.

\begin{table}[H]
  \caption{Components of the sentences in the GRID corpus}
  \label{tab:grid}
  \begin{center}
  \begin{tabular}{|c||c|c|c|}
  \hline
  Category & Content & Number \\
  \hline
  \hline
  Command  &  bin, lay, place, set  &  4\\ 
  \hline 
 Color  &  blue, green, red, white  &  4\\ 
  \hline 
 Preposition  &  at, by, in, with  &  4\\ 
  \hline 
 letter  &  A, B, ... Y, Z (without W)  &  25\\ 
  \hline 
 Digit  &  zero, one, ... eight, nine  &  10\\ 
  \hline 
 Adverb  &  again, now, please, soon  &  4\\ 
  \hline 
  \end{tabular}
  \end{center}
  \end{table}

The original LipNet and our modified LipNet with FPA module are trained 
and evaluated on the GRID corpus. We follow the datasets split setting in \cite{Assael2016} and use two male speakers (speaker 1 and 2) and two female speakers (speaker 20 and 22) for evaluation (3971 videos are usable after filtrating invalid videos). The remaining videos are used for training (28775 videos are usable). By this setting, the models are evaluated on speakers that have not appeared in the training process, thus testing the generalization ability of the models in a more convincing way.  

\subsubsection{Implementation Details}

We use the Keras implementation of LipNet on the GitHub and attain $17.7\%$ word error rate (WER) on the evaluation set, higher than the result of $11.4\% $ in the original paper. The reason may be attributed to not introducing video clips of individual words as additional training instances or other deficient operations in the training process. Since the goal of our work is to test the validity of 3D-FPA, we did not finetine the networks further to achieve the WER declared in \cite{Assael2016}. 

Inside the FPA module, we apply Batch Normalization layers and Dropout layers after each convolutional layer. Since GRID is not very large and the model is easy to overfit, we apply Dropout layers to alleviate this problem. When upsampling the high-level feature maps, there is a problem that the upsampled features don't have the same shape with previous feature maps so that they cannot be added. Then we employ bilinear upsampling if the spatial sizes don't match  and padding the last frame to the end before upsampling if the temporal sizes don't match.

We use Adam as the optimizer and set the learning rate as 0.0001, a first-moment momentum coefficient as 0.9, a second-moment momentum coefficient as 0.999.

\subsubsection{Results and Analysis}

We embed 2D-FPA and 3D-FPA at different position in the original architecture as shown in Figure \ref{fig:exp} (a) (b) (c) respectively, train and evaluate it the same way as baseline method. The results are shown in Table \ref{tab:lipnet}. CER, WER and BLEU are short for character-level error rate, word-level error rate and Bilingual Evaluation Understudy respectively. The ``input'' in the parentheses means that the 3D-FPA is embedded after the input frames, ``F1'' means it is embedded after F1. ``F1, F2'' means we simultaneously apply 3D-FPA modules after F1 and F2. The results indicate that the performance of LipNet get a considerable improvement with the 3D-FPA module embedded. 

\begin{table}[H]
\caption{Results of sentence-level experiments on the GRID corpus}
\label{tab:lipnet}
\begin{center}
\begin{tabular}{|c||c|c|c|}
\hline
Method & CER & WER & BLEU\\
\hline
\hline
LipNet & 0.11192 & 0.17741 & 0.82409\\
\hline
LipNet + 2D-FPA(Input) & 0.08721 & 0.15924 & 0.85144\\
\hline
LipNet + 3D-FPA(Input) & 0.07571 & 0.14543 & 0.86840\\
\hline
LipNet + 2D-FPA(F1)& 0.08983 & 0.16033 & 0.84863\\
\hline
LipNet + 3D-FPA(F1)& 0.08019 & 0.14224 & 0.86725\\
\hline
LipNet + 2D-FPA(F2)& 0.08002 & 0.14660 & 0.86441\\
\hline
LipNet + 3D-FPA(F2)& \textbf{0.07246} & \textbf{0.14178} & \textbf{0.87221}\\
\hline
LipNet + 3D-FPA(F1,F2)& 0.07798 & 0.14199 & 0.86956\\
\hline
\end{tabular}
\end{center}
\end{table}

\begin{figure}[H]
  \centering
  \includegraphics[width=1.0\linewidth]{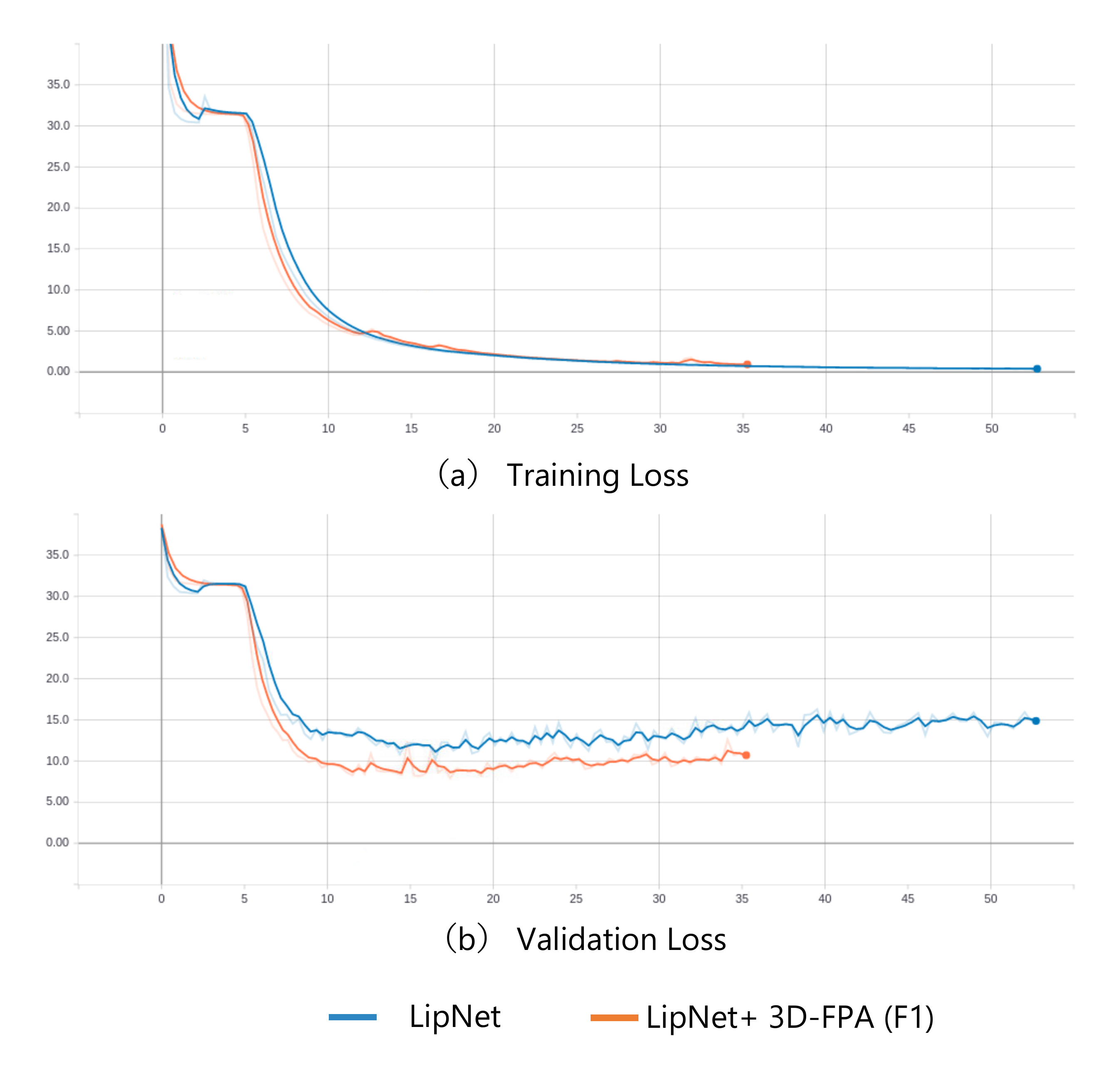}
  \caption{The training loss and validation loss of original LipNet and modified LipNet with 3D-FPA. It indicates that 3D-FPA works well on preventing overfitting.}
  \label{fig:loss}
\end{figure}

Figure \ref{fig:loss} shows the training loss and validation loss during the training process. The training loss of the model with 3D-FPA module starts to decrease earlier than the original model, after a few epochs of stasis. Also, the validation loss of the modified model is lower than the original model, showing that the 3D-FPA module has effect on preventing overfitting.

Moreover, we check the error rate for each word of the sentence. All the sentences in the GRID corpus follow a fixed grammar: command + color + preposition + letter + digit + adverb, for example, ``Set blue by A four please" and ``Place red at C zero again". Table \ref{tab:lipnet_words} shows the performance on each word of the sentences, where the number in the parentheses denotes the sequence number of the word in the sentence. The results show that our method reduces the WER of each word at different degrees. It is noteworthy that it makes great improvement on the fifth and sixth word, inducing more than $50\%$ decline in WER. 

\begin{table*}
  \caption{WER of each word on the GRID corpus}
  \label{tab:lipnet_words}
  \begin{center}
  \begin{tabular}{|c||c|c|c|c|c|c|c|}
  \hline
  Method & WER & WER(1) & WER(2) & WER(3) & WER(4) & WER(5) & WER(6) \\
  \hline
  \hline
  LipNet & 0.17741 &  0.03123 & 0.09166 & 0.21884 & 0.36540 & 0.23218 & 0.12188 \\
  \hline
  LipNet + 2D-FPA(Input) & 0.15924 & \textbf{0.01989} & 0.10073 & 0.24402 & 0.39688 & 0.13624 & 0.05691\\
  \hline
  LipNet + 3D-FPA(Input) & 0.14543 & 0.02115 & 0.10375 & 0.21632 & 0.39612 & 0.07681 & 0.05842\\
  \hline
  LipNet + 2D-FPA(F1) & 0.16033 & 0.02392 & 0.11710 & 0.20071 & 0.41022 & 0.14681 & 0.05943\\
  \hline
  LipNet + 3D-FPA(F1) & 0.14224 & 0.02392 & \textbf{0.09116} & 0.19995 & 0.37094 & 0.10350 & 0.06396\\
  \hline
  LipNet + 2D-FPA(F2)& 0.14660 & 0.04105 & 0.11458 & \textbf{0.19642} & 0.38479 & 0.08965 & 0.05288\\
  \hline
  LipNet + 3D-FPA(F2)& \textbf{0.14178} & 0.02770 & 0.10551 & 0.21355 & 0.38378 & \textbf{0.08084} & \textbf{0.03274}\\
  \hline
  LipNet + 3D-FPA(F1,F2)& 0.14199 & 0.04357 & 0.09771 & 0.19970 & \textbf{0.36439} & 0.08889 & 0.05742\\
  \hline
  \end{tabular}
  \end{center}
  \end{table*}

\subsection{Word-level}

\subsubsection{Dataset}

\begin{figure}[htb]
  \centering
  \includegraphics[width=0.8\linewidth]{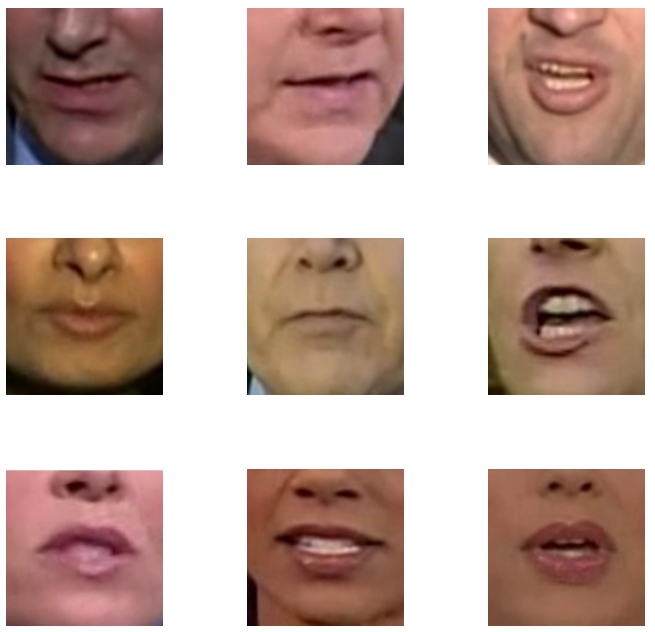}
  \caption{Lip crops in the LRW corpus. Since the videos are extracted from BBC TV broadcasts rather than recorded in laboratory, the data has extraordinarily high variation, making it the most challenging word-level corpus.}
  \label{fig:lrw}
\end{figure}

LRW is the most challenging public word-level database. It contains large amounts of audiovisual speech segments extracted from BBC TV broadcasts. Several characteristics make it challenging. (1) It has 500 target words, compared with 24 words of GRID and 26 words of CUAVE. For each target word, it has a training set of 1000 segments, a validation and an evaluation set of 50 segments each. The total duration of this corpus is 173 hours, compared to 28 hours of GRID. (2) the videos have high variation in the pose, the angle and the age of speakers. Also, the background is set in the wild while most lipreading datasets are recorded indoors in a controlled lab environment. (3) the target words are not isolated in the video segments but appear with a few context words. This add to the difficulty of recognizing the target words. The model is expected to spot on the key frames and ignore the disturbing context. Some cropped frames of LRW are presented in Figure \ref{fig:lrw}.

The model proposed in \cite{Stafylakis2017} and our modified models with FPA embedded are trained and evaluated on LRW. 

\subsubsection{Implementation Details}

We use the PyTorch implementation of CRNL on the GitHub and follow the training instructions in \cite{Stafylakis2017}. Initially , a temporal convolutional back-end is used to replace  the Bi-LSTM. After convergence, the temporal convolution  back-end is removed and the Bi-LSTM back-end is attached. Keeping the weights of the spatiotemporal convolution front-end and the ResNet fixed, the Bi-LSTM back-end is trained for 5 epochs. Finally, the overall systems is trained end-to-end. Our model gets the accuracy of $77.8\%$ on the test set after convergence, lower than the accuracy of $83.0\%$ in the original paper. 

Since LRW is much more complex, we only apply Batch Normalization layers after each convolution layer in the FPA and do not apply Dropout layers.

We employ ResNet-34 network in the architecture and use the standard SGD training algorithm with the initial learning rate of 0.003 the momentum of 0.9. 

\subsubsection{Results and Analysis}

We embed 2D-FPA and 3D-FPA module before and inside ResNet, as shown in Figure \ref{fig:exp2}, import the pretrained weights of the original model and train them on the LRW corpus. The results are shown in Table \ref{tab:crnl}. The performance of CRNL model is improved with FPA modules embedded.

Table \ref{tab:crnl_top} shows 20 words with the highest accuracy predicted by the original CRNL model. Table \ref{tab:crnl_low} shows 10 words with the lowest accuracy predicted by the original model and relative accuracy by modified models. It indicates that the model performs better on longer words, which accords with the fact that human lipreaders perform better on longer words or sentences. Table \ref{tab:crnl_low} also shows that our proposed module improves the performance of the basic architecture on these short words.

We compare the accuracy between the original model and the model with one 3D-FPA moudle embedded shown in Figure \ref{fig:exp2} (a) to find which type of words can be improved more. Table \ref{tab:crnl_growth} shows the words with the highest growth in accuracy, as well as the words that are most likely to be mistaken. The results indicate that our proposed module improves the ability to distinguish words with different suffixes like ``Difference'' and ``Different'', and words with difference in the initial phonemes like ``Million'' and ``Billion'', which is one of the concerns in the lipreading task.

\begin{table}[H]
  \caption{Results of word-level experiments on the LRW corpus}
  \label{tab:crnl}
  \begin{center}
  \begin{tabular}{|c||c|}
  \hline
  Method & Accuracy \\
  \hline
  \hline
  CRNL & 0.7780  \\
  \hline
  CRNL + 2D-FPA &0.7837 \\
  \hline
  CRNL + 3D-FPA*1 & 0.7895 \\
  \hline
  CRNL + 3D-FPA*4 & \textbf{0.7926} \\
  \hline
  \end{tabular}
  \end{center}
  \end{table}

\begin{table}[H]
	\caption{The words with highest accuracy by the original model}
	\label{tab:crnl_top}
	\begin{center}
		\begin{tabular}{|c|c||c|c|}
			\hline
			Ground Truth & Acc & Ground Truth & Acc \\
			\hline
			\hline
      WESTMINSTER  &  1.00  &  INVESTMENT  &  0.98\\ 
      \hline 
     TEMPERATURES  &  1.00  &  CHIEF  &  0.98\\ 
      \hline 
     PRIVATE  &  1.00  &  WOMEN  &  0.98\\ 
      \hline 
     GERMANY  &  1.00  &  MIGRANTS  &  0.98\\ 
      \hline 
     BEFORE  &  1.00  &  TOMORROW  &  0.98\\ 
      \hline 
     WEAPONS  &  1.00  &  FOLLOWING  &  0.96\\ 
      \hline 
     PROVIDE  &  1.00  &  INFORMATION  &  0.96\\ 
      \hline 
     WELFARE  &  1.00  &  PARLIAMENT  &  0.96\\ 
      \hline 
     SUNSHINE  &  0.98  &  POTENTIAL  &  0.96\\ 
      \hline 
     EUROPEAN  &  0.98  &  AFTERNOON  &  0.96\\ 
      \hline 
		\end{tabular}
	\end{center}
\end{table}

  \begin{table}[H]
    \caption{The words with lowest accuracy by the original model and relative accuracy by modified models}
    \label{tab:crnl_low}
    \begin{center}
    \begin{tabular}{|p{1.2cm}||p{1.3cm}|p{1.3cm}|p{1.4cm}|p{1.4cm}|}
    \hline
    Ground Truth  &  Acc (CRNL)  &  Acc (CRNL+ 2D-FPA)  & Acc (CRNL+ 3D-FPA*1)  &  Acc (CRNL+ 3D-FPA*4)\\
    \hline
    \hline
    UNDER  &  0.31  &  \textbf{0.38} &  \textbf{0.38}  &  \textbf{0.38}\\ 
     \hline 
    UNTIL  &  0.35  &  \textbf{0.44}  &  0.38  &  0.42\\ 
     \hline 
    THEIR  &  0.36  &  0.44  &  0.43  &  \textbf{0.48}\\ 
     \hline 
    THERE  &  0.38  &  0.38  &  \textbf{0.44}  &  0.34\\ 
     \hline 
    THINK  &  0.40  &  \textbf{0.48}  &  0.46  &  0.39\\ 
     \hline 
    STAND  &  0.41  &  0.46  &  \textbf{0.52}  &  0.48\\ 
     \hline 
    THING  &  0.42  &  0.46  &  0.44  &  \textbf{0.52}\\ 
     \hline 
    TAKING  &  0.44  &  0.60  &  \textbf{0.60}  &  0.56\\ 
     \hline 
    THESE  &  \textbf{0.46}  &  0.40  &  0.40  &  0.42\\ 
     \hline 
    COULD  &  0.48  &  0.44  &  \textbf{0.52}  &  0.51\\ 
     \hline     
    \end{tabular}
    \end{center}
    \end{table}

    \begin{table}[H]
      \caption{The words with highest growth in accuracy}
      \label{tab:crnl_growth}
      \begin{center}
      \begin{tabular}{|p{1.5cm}||p{0.7cm}|p{1.4cm}|p{0.7cm}|p{1.4cm}|p{0.6cm}|}
      \hline
      Ground Truth & Acc (CRNL)  & Acc (CRNL+  3D-FPA*1)  & Growth  & Confusing Word  & Error Rate \\
      \hline
      \hline
      WORDS  &  0.58  &  0.76  &  0.18  &  WORLD  &  0.10\\ 
      \hline 
     PRICE  &  0.52  &  0.68  &  0.16  &  PRESS  &  0.08\\ 
      \hline 
     DIFFERENCE  &  0.72  &  0.88  &  0.16  &  DIFFERENT  &  0.12\\ 
      \hline 
     TAKING  &  0.44  &  0.60  &  0.16  &  TAKEN  &  0.14\\ 
      \hline 
     BANKS  &  0.64  &  0.80  &  0.16  &  PLACE  &  0.10\\ 
      \hline 
     RUSSIAN  &  0.61  &  0.76  &  0.15  &  RUSSIA  &  0.27\\ 
      \hline 
     LEADERS  &  0.58  &  0.72  &  0.14  &  LEAST  &  0.08\\ 
      \hline 
     CLOSE  &  0.66  &  0.80  &  0.14  &  ALLOWED  &  0.06\\ 
      \hline 
     MORNING  &  0.86  &  0.98  &  0.12  &  POINT  &  0.04\\ 
      \hline 
     MILLION  &  0.64  &  0.76  &  0.12  &  BILLION  &  0.10\\ 
      \hline 
      \end{tabular}
      \end{center}
      \end{table}

\section{Conclusion}

We propose a spatiotemporal pixel-level attention module for visual speech recognition. It utilize high-level features to guide the attention of low-level features, exploring the multi-scale context information in both temporal and spatial domains. Notably, it proposes a creative usage of temporal pyramid by combining pyramidal structures with attention mechanism. Because of its concise structure and low computation cost, 3D-FPA can be embedded into different backbone architectures, therefore has great potential in other Video Understanding tasks besides visual speech recognition.



{\small
\bibliographystyle{ieee}
\bibliography{refs}
}

\end{document}